\documentclass[10pt,twocolumn,letterpaper]{article}

\usepackage{iccv}              
\usepackage{booktabs}
\usepackage{multirow}
\usepackage{graphicx}
\usepackage{indentfirst}
\usepackage[accsupp]{axessibility}
\definecolor{iccvblue}{rgb}{0.21,0.49,0.74}
\usepackage[pagebackref,breaklinks,colorlinks,allcolors=iccvblue]{hyperref}

\def\paperID{5311} 
\def\confName{ICCV}
\def\confYear{2025}

\title{LLM-Assisted Semantic Guidance for Sparsely Annotated Remote Sensing Object Detection}

\author{
Wei Liao$^1$, Chunyan Xu$^{1}$\thanks{Corresponding Author: C. Xu (cyx@njust.edu.cn).}, Chenxu Wang$^1$, Zhen Cui$^{2}$\\
$^1$Nanjing University of Science and Technology, Nanjing, Jiangsu, China\\
$^2$Beijing Normal University, Beijing, China
}

\begin{document}
\maketitle
\begin{abstract}
\indent Sparse annotation in remote sensing object detection poses significant challenges due to dense object distributions and category imbalances. 
Although existing Dense Pseudo-Label methods have demonstrated substantial potential in pseudo-labeling tasks, they remain constrained by selection ambiguities and inconsistencies in confidence estimation.
In this paper, we introduce an LLM-assisted semantic guidance framework tailored for sparsely annotated remote sensing object detection, exploiting the advanced semantic reasoning capabilities of large language models (LLMs) to distill high-confidence pseudo-labels.
By integrating LLM-generated semantic priors, we propose a Class-Aware Dense Pseudo-Label Assignment mechanism that adaptively assigns pseudo-labels for both unlabeled and sparsely labeled data, ensuring robust supervision across varying data distributions. 
Additionally, we develop an Adaptive Hard-Negative Reweighting Module to stabilize the supervised learning branch by mitigating the influence of confounding background information.
Extensive experiments on DOTA and HRSC2016 demonstrate that the proposed method outperforms existing single-stage detector-based frameworks, significantly improving detection performance under sparse annotations. Our source code is available at \url{https://github.com/wuxiuzhilianni/RSST}.
\end{abstract}

\section{Introduction}

\begin{figure}[t]
  \centering
  \includegraphics[width=0.475\textwidth]{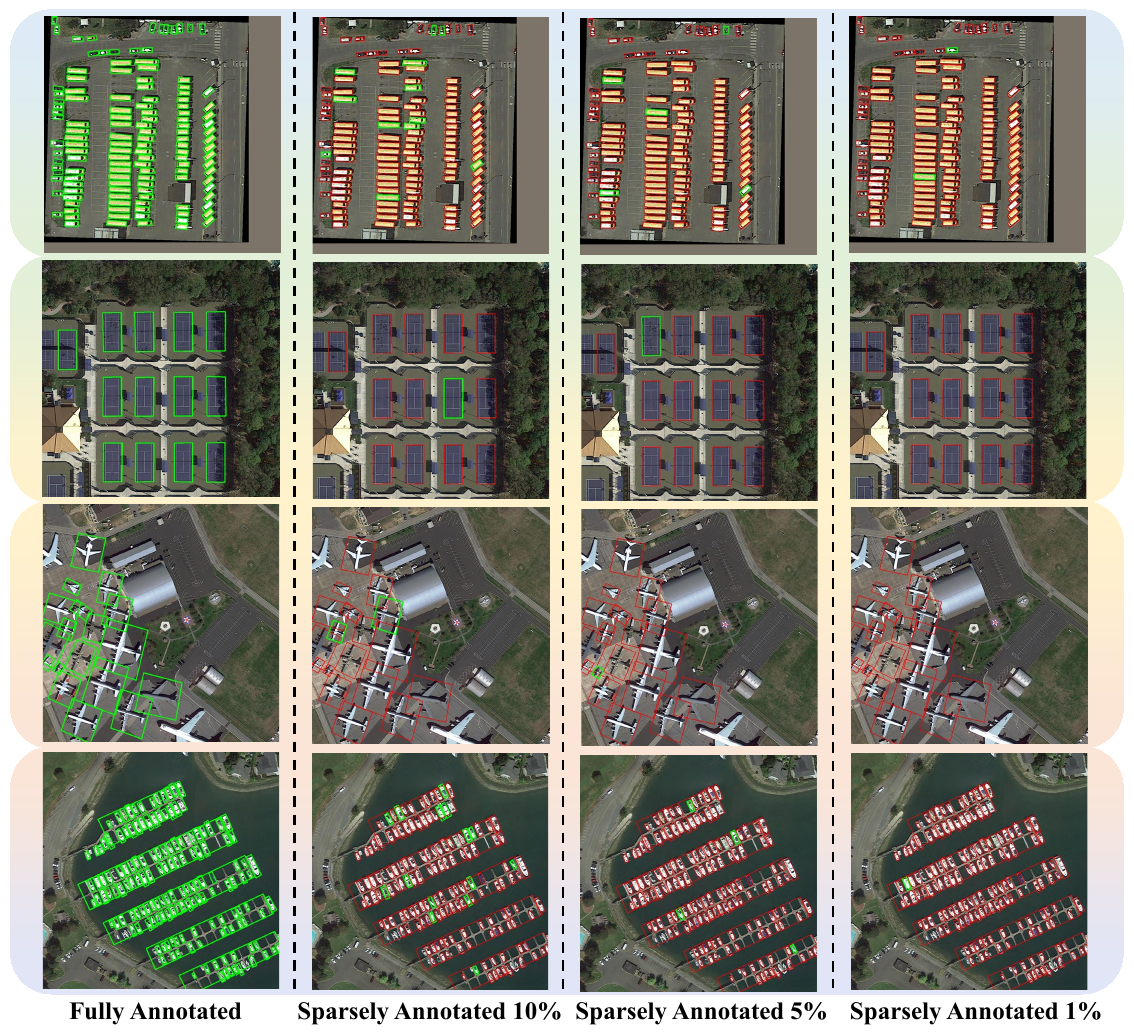} 
  \vspace{-0.5cm}
  \caption{The illustration of sparsely annotated objects in aerial images. From left to right, the columns represent the fully annotated setting, as well as sparsely annotated settings with 10\%, 5\%, and 1\% of the original annotations, respectively. Green bounding boxes indicate the ground truth annotations in the original dataset, while red bounding boxes denote annotations that have been intentionally removed during training.} \vspace{-0.6cm}
  \label{fig:show_dataset}
\end{figure}

\indent Object detection has been a pivotal task in computer vision, underpinning a wide spectrum of real-world applications, including autonomous driving, surveillance, and remote sensing analysis. Traditional object detection models require extensive human-annotated datasets to ensure optimal performance \cite{COCO, Pascal}. To mitigate this reliance on fully annotated data, \textit{semi-supervised object detection} (SSOD) \cite{SOOD++, chen2022label, tang2021proposal} has emerged as a promising paradigm, leveraging a limited set of fully labeled images alongside an abundant corpus of unlabeled images to enhance model generalization while alleviating annotation burdens.  
Despite its success in generic vision tasks, SSOD remains underexplored in the domain of remote sensing imagery \cite{SOOD, Pseco, Consistent-teacher}. This discrepancy stems from the intrinsic challenges posed by remote sensing data, wherein objects are often densely distributed, exhibit significant intra-class variations, and manifest diverse orientations across images. To address this limitation, researchers have introduced the concept of \textit{sparsely annotated object detection} (SAOD) \cite{PECL, rambhatla2022sparsely, Sparsedet}, where only a fraction of objects within an image are explicitly labeled, while the rest remain unannotated. Fig.~\ref{fig:show_dataset} presents a schematic illustration of sparse annotations with varying label rates for remote sensing images.

\indent Pseudo-labeling \cite{Pseudo-label}, in conjunction with consistency-based regularization \cite{Consistency-based-regularization}, forms a cornerstone of SSOD. In this paradigm, the teacher detector generates pseudo-labels, which serve as supervisory signals for training the student model on unlabeled data, thereby reducing reliance on manually annotated datasets. Some methods adopt pseudo-boxes, where the predicted bounding boxes undergo multiple refinement procedures and serve as direct supervision for the student detector \cite{SoftTeacher, UnbiasedTeacher, UnbiasedTeacherV2}. However, these additional post-processing steps introduce potential information loss and may lead to suboptimal supervision quality. Alternatively, other approaches employ Dense Pseudo-Labels (DPL) by leveraging raw model outputs without conventional post-processing \cite{DenseTeacher, ARSL, MCL}, which can provide richer supervision signals and potentially enhance detection performance under limited annotations \cite{Dtg-ssod}.

\indent Prior methodologies leveraging DPL predominantly employ a top-$k$ selection mechanism to identify foreground pixels for distillation. However, this heuristic approach introduces inherent selection and assignment ambiguities, which undermine the effectiveness of pseudo-labeling. To alleviate these ambiguities, recent works have explored alternative selection paradigms. For instance, ARSL \cite{ARSL} refines pseudo-label selection by jointly considering classification confidence and localization quality, whereas MCL \cite{MCL} adopts a divide-and-rule strategy, wherein distinct selection rules are formulated for objects of varying scales. Despite these improvements, a fundamental inconsistency arises between the mean confidence of pseudo-labels and their associated categorical assignments. Specifically, some underrepresented categories exhibit relatively elevated confidence scores, leading to their frequent selection during the assignment process. In contrast, categories with larger sample sizes tend to display markedly lower confidence, resulting in inadequate model training. This misalignment between localized confidence estimates and the true underlying semantic structures introduces substantial ambiguity, thereby disrupting the model's learning dynamics.

\indent To address the aforementioned challenges, we propose an LLM-assisted semantic guidance framework grounded in a Multi-Branch Input (MBI) architecture, which synergistically integrates supervised and unsupervised learning paradigms to enhance semantic consistency and pseudo-label reliability. Within the supervised learning branch, the Adaptive Hard-Negative Reweighting (AHR) module is designed to dynamically modulate the influence of hard negatives in the logit space, thereby mitigating their adverse effects on model optimization. Concurrently, in the unsupervised learning branch, the LLM-Assisted Semantic Prediction (LSP) module autonomously generates and gradually refines class-specific prompts, ensuring the progressive enhancement of semantic representations. These prompts subsequently support the Class-Aware Label Assignment (CLA) mechanism, which adaptively modulates the assignment strategies to accommodate both fully unlabeled and sparsely labeled data distributions. By incorporating these synergistic components, the proposed framework effectively mitigates the misalignment between localized confidence estimates and the underlying semantic structures, ultimately fostering a more robust and semantically coherent learning paradigm.


\indent To summarize, the principal contributions of this work are as follows:
    i) We propose an LLM-assisted semantic guidance framework, meticulously tailored to accommodate the challenges posed by the sparsely annotated object detection (SAOD) task. 
    ii) To alleviate the misalignment between localized confidence estimates and the underlying semantic structures, we have constructed several effective modules for both the supervised and unsupervised branches.
    iii) We conduct comprehensive experiments on two remote sensing datasets, including DOTA \cite{DOTA} and HRSC2016 \cite{HRSC2016}, demonstrating the effectiveness and robustness of our model.

\vspace{-0.2cm}
\section{Related Work}
\label{sec:formatting}
\subsection{Oriented Object Detection}

\indent Oriented object detection has emerged as a vital research area in computer vision, addressing the challenges posed by objects with arbitrary orientations, particularly in aerial image analysis. Unlike traditional object detectors that rely on horizontal bounding boxes (HBB) \cite{FastRCNN, FasterRCNN, YOLO, FCOS, zhao2024weakly}, oriented object detection techniques utilize oriented bounding boxes (OBB), which provide a more accurate representation of rotated objects \cite{ReDet, R3Det, ORCNN}. 
In recent years, several innovative approaches have been developed to improve the accuracy and robustness of oriented object detection. CSL \cite{CSL} reformulates the angle regression problem as a classification task to address the issue of discontinuous boundaries caused by angular periodicity or corner ordering, introducing a circular smooth label technique to improve error tolerance and enhance detection performance.
S2A-Net \cite{S2Anet} leverages the feature alignment module to generate high-quality anchors, followed by the oriented detection module, which employs active rotating filters to produce features that are both orientation-sensitive and orientation-invariant. 
In contrast to the above studies, which primarily focus on the supervised learning paradigm, this work takes an initial step towards exploring sparsely annotated oriented object detection, aiming to reduce annotation costs and improve detector performance by utilizing unlabeled data.

\subsection{Label-Efficient Object Detection}

\indent Label-efficient object detection aims to mitigate the dependency on fully annotated datasets by leveraging weak supervision, self-training, and pseudo-labeling techniques. Semi-supervised object detection (SSOD) has advanced with teacher-student frameworks, where the teacher model, updated via exponential moving average (EMA), generates pseudo-labels to guide student training \cite{zhang2022semi, mixteacher, wu2024pseudo}. Soft Teacher \cite{SoftTeacher} employs box jittering for pseudo-box refinement, while Unbiased Teacher \cite{UnbiasedTeacher} and its improved variant \cite{UnbiasedTeacherV2} address class imbalance and enhance anchor-free detectors, respectively. Dense Teacher \cite{DenseTeacher} introduces dense pseudo-labels to replace heuristic post-processing steps. ARSL \cite{ARSL} integrates joint-confidence estimation and task-separation assignment to improve label assignment at the pixel level. Sparsely annotated object detection (SAOD) addresses scenarios where only a fraction of instances are labeled, reducing annotation costs while maintaining detection performance. The absence of complete annotations introduces challenges, such as missing supervision signals. Niitani et al. \cite{Niitani_2019_CVPR} propose part-aware sampling to mitigate incorrect supervision. Co-mining \cite{Co-mining} employs a Siamese network to enhance multi-view learning and better mine unlabeled instances. Region-based approaches \cite{rambhatla2022sparsely} treat SAOD as a semi-supervised problem, identifying unlabeled foreground regions. Calibrated Teacher \cite{CalibratedTeacher} refines confidence estimation to stabilize training, ensuring consistent pseudo-label quality. PECL \cite{PECL} introduces a progressive selection strategy tailored to aerial images. In contrast to prior methods, our approach innovatively incorporates dense label representations into sparsely annotated aerial object detection, refining optimization under minimal supervision by integrating self-training and leveraging interclass relational priors to enhance detection robustness.

\section{Method}

\begin{figure*}[t]
  \centering
  \includegraphics[width=\textwidth]{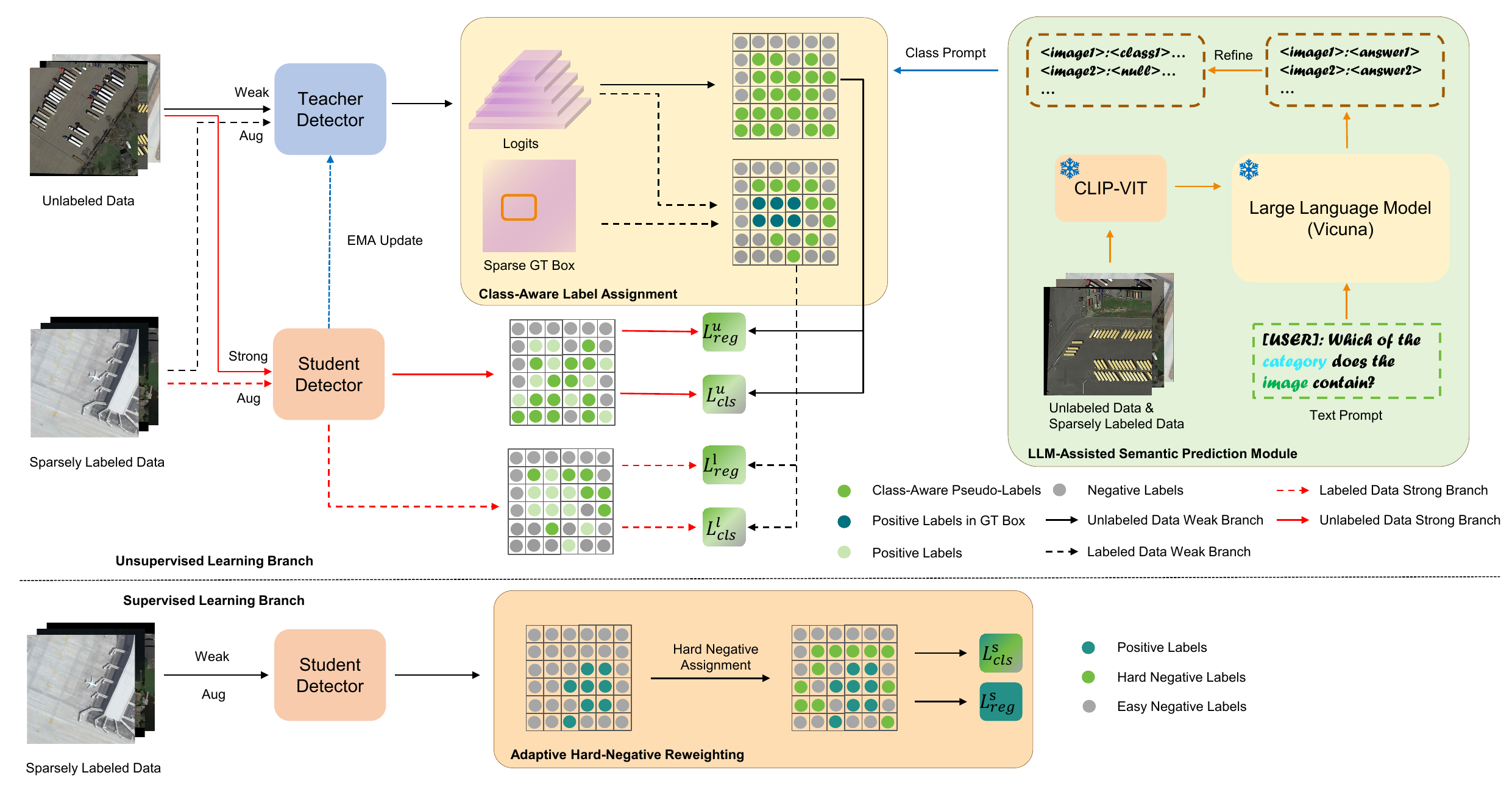} 
  \vspace{-0.6cm}
  \caption{\indent The illustration of the LLM-assisted semantic guidance framework. In the unsupervised branch, the LLM-Assisted Semantic Prediction (LSP) module generates and refines class prompts, while the Class-Aware Label Assignment (CLA) selects high-quality pixel-level pseudo-labels using distinct strategies for sparsely labeled and unlabeled data. In the supervised branch, the Adaptive Hard-Negative Reweighting (AHR) module identifies hard negatives in the logit space and mitigates their impact by reducing weights.} 
  \vspace{-0.4cm}
  \label{fig:overall_framework}
\end{figure*}

\subsection{Overview}

\indent SAOD presents significant challenges due to limited supervisory signals, necessitating the effective utilization of sparsely annotated instances and unlabeled images to train a robust detector capable of accurately identifying all valid targets. Formally, the sparsely annotated training set is defined as \( \mathcal{X}_t = \{ (x_i, y_i) \}_{i=1}^{N} \), where \( N \) represents the total number of images in the training dataset. For each image, denoted as the \( i \)-th sample, its sparse annotations are represented as \( y_i = \{ (c_i^j, \theta_i^j, b_i^j) \}_{j=1}^{N_{il}} \), where \( c_i^j \), \( \theta_i^j \), and \( b_i^j \) correspond to the class label, orientation, and bounding box coordinates of the \( j \)-th annotated object, respectively. In this context, \( N_{il} \) denotes the number of labeled instances present in the \( i \)-th image, with \( N_{il} \geq 0 \) indicating the possibility of zero labeled instances.

\indent Fig.~\ref{fig:overall_framework} presents an overview of the LLM-assisted semantic guidance framework. The framework is designed as a self-supervised, single-stage detection paradigm based on the Multi-Branch Input (MBI) Architecture, where a tightly coupled, cyclic optimization mechanism coordinates the reciprocal refinement between the generation of conformal dense pseudo-labels and the progressive adaptation of the detector. The framework comprises a supervised and an unsupervised branch, each playing a distinct role in optimizing the model's learning process.
In the supervised branch, the model is trained with sparsely annotated data, providing the primary supervisory signal that guides the optimization trajectory of the student detector throughout the training process. During this phase, the Adaptive Hard-Negative Reweighting (AHR) module is employed to regulate hard negatives, facilitating a more discriminative learning process.
In the unsupervised branch, the LLM-Assisted Semantic Prediction (LSP) module utilizes a large language model (LLM) to autonomously generate and progressively refine class-specific prompts for each image in an offline manner. This refinement significantly enhances the subsequent Class-Aware Label Assignment (CLA) mechanism, which strategically adapts the assignment strategies based on both labeled and sparsely labeled data. This approach mitigates foreground class imbalance and strengthens the overall robustness of the detection framework.

\subsection{LLM-Assisted Semantic Prediction}



\begin{figure}[t]
  \centering
  \includegraphics[width=0.475\textwidth]{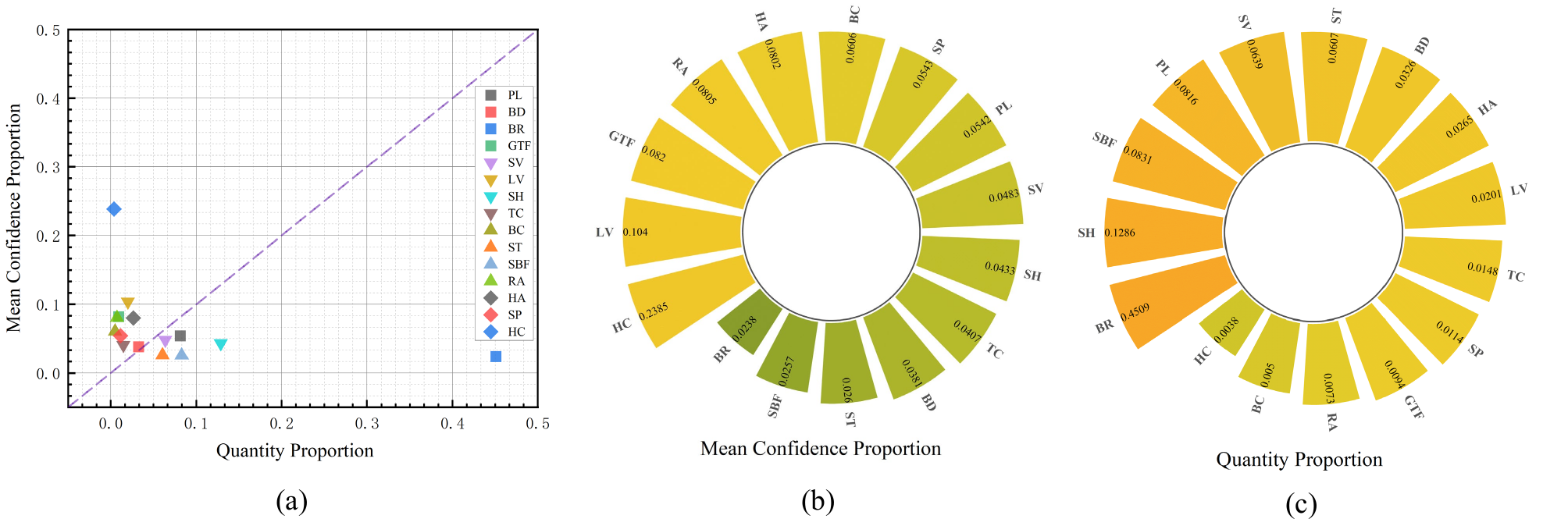} \vspace{-0.6cm}
  \caption{Analysis of Imbalance Between Category-Wise Quantity Proportion and Mean Confidence Proportion. (a) A scatter plot illustrating the relationship between Quantity Proportion (x-axis) and Mean Confidence Proportion (y-axis) for each category. (b) A radial bar chart displaying the Quantity Proportion of each category in ascending order, with category labels annotated accordingly. (c) A radial bar chart presenting the Mean Confidence Proportion of each category, sorted in ascending order, with category labels provided.}
  \vspace{-0.4cm}
  \label{fig:num_confidence}
\end{figure}

\indent Prevailing strategies employing DPL for model training primarily delineate label assignment protocols by identifying the top-$k$ most confident predictions either across the global spatial extent~\cite{DenseTeacher}, or within heterogeneous feature maps through a divide-and-rule paradigm~\cite{MCL}.
However, such approaches intrinsically overlook a pivotal aspect: the explicit semantic determination of foreground categories. As illustrated in Fig.~\ref{fig:num_confidence}, certain underrepresented categories exhibit relatively high confidence scores, leading to their frequent selection during the assignment process. Conversely, some categories with a larger sample count tend to have significantly lower confidence scores, resulting in insufficient training for the model. The misalignment between localized confidence estimations and the true underlying semantic structures introduces substantial ambiguity into the learning dynamics.

\indent Drawing inspiration from \cite{GeoChat}, we address this challenge by employing a large language model (LLM) to assist offline pseudo-label assignment before the training stage. 
For each image, the LLM is leveraged to infer potential foreground categories, which are subsequently utilized to enhance the unsupervised label assignment. 
Specifically, given the complete set of categories in the dataset, denoted as \( \mathcal{C} = \{\textit{class}_1, \textit{class}_2, \dots, \textit{class}_{15}\} \) in DOTA, along with an additional background category \( \varnothing \), we formulate a structured linguistic query to guide the LLM in foreground category inference. The designed instruction is as follows:  
``\texttt{Choose categories presented in the image: class$_1$, ..., class$_{15}$, none. Choose one or several classes. Answer in one word or a short phrase.}"
Next, high-resolution remote sensing imagery is processed through CLIP-ViT (L-14) \cite{CLIP}, where the positional encoding is interpolated to accommodate an expanded input resolution of \( 504 \times 504 \), increasing the number of patches to 1296. An MLP-based adaptor then maps the 1024-dimensional visual embeddings into a 4096-dimensional space, ensuring compatibility with the Vicuna-v1.5 (7B) model \cite{Vicuna}. The structured visual representations, concatenated with the formulated query, are fed into the LLM to facilitate semantically coherent category identification. Leveraging its contextual reasoning within an autoregressive decoding paradigm, the LLM derives the most probable foreground categories. If the model does not identify any foreground objects, it outputs \texttt{none}.  
Once foreground predictions are obtained for all images, a refinement step is applied: for sparsely annotated data, where labeled instances already contain foreground category information, the final category set is derived by taking the union of the LLM-generated predictions and the provided annotations. In contrast, for unlabeled data, the predictions remain unchanged. This refinement strategy ultimately ensures that approximately four-fifths of the final predictions are correct.

\subsection{Class-Aware Label Assignment}
    
\indent With the assistance of the class prompt generated by LSP, foreground category information is explicitly leveraged to refine the assignment paradigm of positive and negative samples within the unsupervised learning distillation process. By embedding class-specific priors into the distillation framework, the proposed methodology mitigates the intrinsic ambiguities associated with high-confidence pseudo-label predictions, thereby enhancing their semantic reliability and robustness. To  optimize the selection of supervisory signals, distinct assignment strategies are employed for different data types.
To further handle fully unlabeled data, we propose a label assignment method that leverages the class prompt generated by the LSP module. This method aims to select a fixed number of foreground pixels through a structured three-stage process, ensuring that the selected pixels exhibit high semantic relevance and are well-distributed across different foreground categories.

\indent Given a pixel set \( P = \{p_i\} \), we prioritize selecting pixels that are semantically associated with foreground categories, as specified by a predefined class prompt. This ensures that the selected pixels are aligned with the intended semantic content. To further enhance the reliability of the selection process and suppress potentially noisy or ambiguous samples, pixels with confidence scores falling below a predefined threshold \( thr \) are systematically filtered out:
\begin{equation}
P_{\text{fg}} = \{ p \in P \mid C(p) \in \text{Prompt} \ \wedge \ S(p) > thr \},
\end{equation}
where \( C(p) \) denotes the predicted category of pixel \( p \), and \( S(p) \) represents the joint confidence score of pixel \( p \), as defined in \cite{ARSL}. The threshold \( thr \) serves as a confidence filter, removing ambiguous pixels with the low reliability.
To enhance generalization, we introduce an additional selection mechanism based on the pixel confidence. Specifically, we select the top-\( k \) pixels with the highest joint confidence scores:
\begin{equation}
P_{\text{conf}} = \operatorname{Top}_k ( \sum_{p \in P} S(p) ),
\end{equation}
where \( \operatorname{Top}_k(\cdot) \) denotes the operation of selecting the highest-\( k \) scoring pixels.
To ensure balanced representation across different foreground categories, we further refine the pixel selection by choosing the top-\( k_j \) pixels with the highest confidence scores within each individual category \( c_j \):
\begin{equation}
P_{c_j} = \operatorname{Top}_{k_j} ( \sum_{p \in P} S_j(p) ),
\end{equation}
where \( S_j(p) \) represents the confidence measure of the pixel \( p \) corresponding to the category \( c_j \). Ultimately, to eliminate redundancy and enhance the heterogeneity of the selected pixels, a deduplication operation is conducted. The resulting final set of selected pixels from the unlabeled images is defined as follows:
\begin{equation}
P_{\text{unlabeled}} = \operatorname{Unique} \big( P_{\text{fg}} \cup P_{\text{conf}} \cup \bigcup_{j} P_{c_j} \big),
\end{equation}
where \( \operatorname{Unique}(\cdot) \) ensures that each selected pixel appears only once in the final set. This final selection process guarantees a diverse and semantically meaningful assignment of pixels, facilitating more robust learning.

\indent To address sparsely labeled data, we propose an assignment strategy that incorporates additional supervisory signals from available annotations, including both the class prompts generated by the LSP module and the ground truth information. Given the set of annotated pixels \( P_{\text{GT}} \), we directly include these pixels as positive samples:

\begin{equation}
P_{\text{GT}} = \{ p \in P \mid p \in \mathcal{A} \},
\end{equation}
where $\mathcal{A}$ represents the set of manually labeled pixels. By incorporating these reliable ground-truth (GT) pixels, we ensure that the model receives explicit supervision from human-labeled data, thereby mitigating potential errors introduced by the pseudo-labeling process. The final selection of positive samples combines four sources: (1) pixels aligned with the class prompt, (2) high-confidence pixels, (3) category-balanced pixels, and (4) ground-truth pixels. To eliminate redundancy and maintain sample diversity, a deduplication process is applied:
\begin{equation}
P_{\text{sparsely}} = \operatorname{Unique} ( P_{\text{fg}} \cup P_{\text{conf}} \cup \bigcup_{j} P_{c_j} \cup P_{\text{GT}} ),
\end{equation}
where \( \operatorname{Unique}(\cdot) \) removes duplicate entries, ensuring that each selected pixel appears only once in the final set. This strategy effectively enhances the supervision quality, enabling more reliable training under sparse annotations.

\subsection{Adaptive Hard-Negative Reweighting}


\indent One-stage detectors are inherently more vulnerable to incomplete or missing annotations \cite{BRL, Generalized_focal_loss, Focal_loss}, particularly in the SAOD task, where distinguishing hard negative samples poses a significant challenge. These hard negatives may correspond to genuinely unannotated objects or background clutter, leading to increased uncertainty and hindering model convergence.

\indent To address this challenge, we propose AHR to adaptively modulate the loss contribution of negative samples based on their confidence scores. This strategy aims to strike a balance between mitigating the impact of erroneous supervision and preserving beneficial background cues. The loss function is formulated as follows:  
\begin{equation}
L(p_t) =
\begin{cases} 
- \log(p_t) \alpha p_t^{\gamma},  \\
- \log(1 - p_t) (1 - \alpha) (1 - p_t)^{\gamma},  \\
- \log(1 - p_t) (1 - \alpha) (1 - p_t)^{\gamma} w,  \\
\end{cases}
\end{equation}
where \( p_t \) represents the predicted confidence score, \( \alpha \) controls the weighting of positive and negative samples, and \( \gamma \) adjusts the focusing effect to suppress well-classified samples. The adaptive scaling factor \( w \) down-weights misleading high-confidence negatives. The three cases in the loss function correspond to:  positive samples,  negative samples with confidence below a threshold \( \textit{thr} \), and  negative samples with confidence above \( \textit{thr} \).

\section{Experiments}

\subsection{Experimental Setup}

\textbf{Datasets:} We evaluate our approach on two widely used aerial object detection benchmarks: DOTA \cite{DOTA} and HRSC2016 \cite{HRSC2016}. The DOTA dataset consists of 2,806 high-resolution aerial images with 188,282 annotated instances exhibiting significant variations in scale, aspect ratio, and orientation.The dataset is partitioned into three subsets: a training set containing 1411 images, a validation set with 458 images, and a test set comprising 937 images. It encompasses 15 distinct object categories, namely plane (PL), baseball-diamond (BD), bridge (BR), ground-track-field (GTF), small-vehicle (SV), large-vehicle (LV), ship (SH), soccer-ball field (SBF), tennis-court (TC), basketball-court (BC), storage-tank (ST), roundabout (RA), harbor (HA), swimming-pool (SP), and helicopter (HC). 
Under sparse annotation settings, we randomly sample 1\%, 2\%, 5\%, and 10\% of instances per category before tiling each image into non-overlapping $1024 \times 1024$ patches. The HRSC2016 dataset contains 436 training, 181 validation, and 444 test images. Following \cite{PECL}, we construct sparsely labeled variants by retaining 10\% of annotations, with at least one per category per image. The performance is evaluated using the standard mean average precision (mAP).

\textbf{Implementation Details:} Our proposed models are implemented and trained using the MMDetection \cite{mmdetection} and MMRotate \cite{mmrotate} frameworks. Within this framework, we employ the Rotated FCOS \cite{FCOS} as the single-stage detection paradigm, leveraging a ResNet-50 \cite{ResNet} backbone pretrained on ImageNet \cite{ImageNet} to extract hierarchical feature representations. We employ Stochastic Gradient Descent (SGD) as the optimization algorithm, with an initial learning rate of 0.0025, a momentum coefficient of 0.9, and a weight decay factor of 0.0001 to prevent overfitting.  The total training schedule comprises 12,000 iterations, with an initial burn-in phase of 6,400 iterations to stabilize model convergence. 

\begin{table}[htbp]
\centering
\scriptsize
\caption{Performance comparison of the OBB task under different annotation rates (1\%, 2\%, 5\%, and 10\%) on the DOTA dataset. Only the main mAP results are reported. The detection methods are based on Rotated-FCOS ($\dag$), Rotated-RetinaNet ($\ddag$) and Faster-RCNN ($\circ$).}
\label{tab:map_rate_comparison}
\renewcommand{\arraystretch}{1.2}
\setlength{\tabcolsep}{4pt}
\vspace{-0.2cm}
\begin{tabular}{c l c c c c}
    \specialrule{1.2pt}{0pt}{0pt}
    \textbf{Model Type} & \textbf{Model Name} & \textbf{1\%} & \textbf{2\%} & \textbf{5\%} & \textbf{10\%} \\
    \midrule
    \multirow{6}{*}{Supervised}
        & Rotated FCOS & 41.18 & 43.46 & 48.18 & 49.79 \\
        & Rotated RetinaNet & 43.12 & 45.09 & 48.86 & 53.16 \\
        & S$^2$A-Net & 40.14 & 41.92 & 48.78 & 54.39 \\
        & Rotated Faster R-CNN & 45.44 & 48.31 & 51.81 & 57.75 \\
        & Oriented R-CNN & 51.05 & 53.09 & 56.85 & 61.50 \\
        & ReDet & 50.72 & 52.08 & 58.01 & 61.79 \\
    \midrule
    \multirow{6}{*}{Semi-Supervised}
        & Unbiased Teacher$^{\circ}$ & 43.59 & 44.97 & 51.55 & 57.23 \\
        & Dense Teacher$^{\dag}$ & - & 47.72 & 50.82 & 58.13 \\
        & PseCo$^{\circ}$ & 43.03 & 46.10 & 52.27 & 57.76 \\
        & SOOD$^{\ddag}$ & 45.34 & 47.60 & 52.84 & 57.36 \\
        & ARSL$^{\dag}$ & 43.88 & 44.39 & 51.38 & 55.61 \\
        & MCL$^{\dag}$ & 41.90 & 43.75 & 51.53 & 56.17 \\
    \midrule
    \multirow{2}{*}{Sparsely-Annotated}
        & S$^2$A-Net w/PECL$^{\dag}$ & 50.39 & 53.81 & 57.42 & 62.49 \\
        & Ours$^{\dag}$ & 56.64 & 59.37 & 63.95 & 65.11 \\
    \specialrule{1.2pt}{0pt}{0pt}
    \end{tabular}
    \label{tab:map_comparison}
    \vspace{-0.3cm}
\end{table}

\begin{table}[htbp]
    \centering
    \footnotesize
    \caption{Prediction Statistics of LLM-Assisted Semantic Prediction Module. The total number of images is 21,046.``None" indicates that both the prediction and ground truth (GT) have no foreground classes. ``Exact" means the predicted foreground classes are identical to the GT. ``Partly" denotes cases where the predicted classes are a subset of the GT.}
    \renewcommand{\arraystretch}{1.2}
    \setlength{\tabcolsep}{4pt}
    \vspace{-0.2cm}
    \begin{tabular*}{\linewidth}{@{\extracolsep{\fill}}ccccc}
        \specialrule{1.2pt}{0pt}{0pt}
        \multirow{2}{*}{\textbf{Modification (\%)}} & \multicolumn{3}{c}{\textbf{Correct Predictions}} & \multirow{2}{*}{\textbf{Errors}} \\
        \cline{2-4}  
        & \textbf{None} & \textbf{Exact} & \textbf{Partly} & \\
        \midrule
        0 (Original) & 6943 & 4861 & 3771 & 5471 \\
        1 & 6819 & 7539 & 2259 & 4429 \\
        2 & 6821 & 7755 & 2139 & 4331 \\
        5 & 6829 & 8075 & 1976 & 4166 \\
        10 & 6817 & 8447 & 1765 & 4017 \\
        \specialrule{1.2pt}{0pt}{0pt}
    \end{tabular*}
    \label{tab:prediction_statistics}
    \vspace{-0.3cm}
\end{table}

\begin{table}[htbp]
    \centering
    \scriptsize
    \caption{Prediction Metrics and Cost for Class Prompts Synthesized by Distinct Large Language Models.}
    \renewcommand{\arraystretch}{1.2}
    \setlength{\tabcolsep}{4pt}
    \vspace{-0.2cm}
    \begin{tabular}{c c c c c @{\hspace{1.5pt}} c @{\hspace{1.5pt}} c}
        \specialrule{1.2pt}{0pt}{0pt}
        \textbf{Model} & \textbf{None} & \textbf{Exact} & \textbf{Partly} & \textbf{Errors} & \textbf{Params(B)} & \textbf{T/img (s)} \\
        \midrule
        Phi-3.5-Vision-Instruct    & 2096 & 3036 & 3015 & 12899 & 4.15 & 0.85 \\
        Qwen-VL-Chat-Int4 & 2933 & 2292 & 2097 & 13724 & 3.18 & 0.82 \\
        InternVL2.5-1B    & 2697 & 3177 & 3479 & 11693 & 0.94 & 0.39 \\
        InternVL2.5-2B    & 1752 & 3906 & 3077 & 12311 & 2.21 & 0.44 \\
        InternVL2.5-4B    & 2169 & 4045 & 3506 & 11326 & 3.71 & 0.61 \\
        LLaVA-v1.5-7B     & 4485 & 4182 & 3446 & 8933 & 7.06 & 0.26 \\
        Ours              & 6943 & 4861 & 3771 & 5471 & 7.06 & 0.48 \\
        \specialrule{1.2pt}{0pt}{0pt}
    \end{tabular}
    \label{tab:different_llm}
    \vspace{-0.3cm}
\end{table}

\begin{table}[htbp]
    \centering
    \footnotesize
    \caption{Comparative analysis of prediction consistency under diverse textual prompt formulations within the LLM-Assisted Semantic Prediction module.}
    \renewcommand{\arraystretch}{1.2}
    \setlength{\tabcolsep}{12pt}
    \vspace{-0.2cm}
    \begin{tabular}{c c c c c}
        \specialrule{1.2pt}{0pt}{0pt}
        \textbf{Prompt ID} & \textbf{None} & \textbf{Exact} & \textbf{Partly} & \textbf{Errors} \\
        \midrule
        P1 & - & 682 & 203 & 20161 \\
        P2 & - & 5711 & 4332 & 11003 \\
        P3 & 1224 & 3560 & 3093 & 13169 \\
        P4 & 6979 & 4616 & 3794 & 5657 \\
        P5 & 6943 & 4861 & 3771 & 5471 \\
        \specialrule{1.2pt}{0pt}{0pt}
    \end{tabular}
    \label{tab:different_prompt}
    \begin{minipage}{\linewidth}
    \scriptsize
    \vspace{0.1cm}
    \textbf{Prompt Descriptions:} \\
    P1: \textit{Which of the following categories does the image contain: plane, ..., none.} \\
    P2: \textit{Which of the following categories does the image contain: plane, ..., helicopter. Answer in one word or a short phrase.}\\
    P3: \textit{Please classify this image among: plane, ... , none.} \\
    P4: \textit{What objects are in this image? Choose from the following: plane, ..., none. Answer in one word or a short phrase.} \\
    P5: \textit{Which of the following categories does the image contain: plane, ..., none. Answer in one word or a short phrase.} 
    \end{minipage}
    \vspace{-0.3cm}
\end{table}

\begin{table}[htbp]
    \centering
    \footnotesize
    \caption{Quantitative comparison of representative oriented object detection methods on the HRSC2016 dataset under the Oriented Bounding Box (OBB) setting. The performance is evaluated in terms of mAP, as well as AP at 50\% and 75\% IoU thresholds. }
    \renewcommand{\arraystretch}{1.2}
    \vspace{-0.2cm}
    \begin{tabular*}{\linewidth}{@{\extracolsep{\fill}}c c c c}  
        \specialrule{1.2pt}{0pt}{0pt}
        \multirow{2}{*}{\textbf{Method}} & \multicolumn{3}{c}{\textbf{Performance (\%) $\uparrow$}} \\
        \cmidrule(ll){2-4}
        & \textbf{mAP} & \textbf{AP50} & \textbf{AP75} \\
        \midrule
        Rotated RetinaNet & 47.96 & 82.70 & 49.90 \\
        S$^2$A-Net & 48.05 & 79.60 & 51.60 \\
        Rotated FCOS & 49.85 & 79.70 & 56.10 \\
        Oriented RCNN & 57.26 & 80.90 & 68.70 \\
        ReDet & 63.80 & 88.50 & 77.50 \\
        Dense Teacher & 64.86 & 88.00 & 75.60 \\
        Ours & \textbf{66.80} & 88.40 & 77.40 \\
        \specialrule{1.2pt}{0pt}{0pt}
    \end{tabular*}
    \label{tab:hrsc_comparison}
    \vspace{-0.3cm}
\end{table}

\subsection{Experimental Results}

\indent \textbf{DOTA:} 
We report the performance of our method on the oriented bounding box (OBB) detection task using the DOTA benchmark, as detailed in Table~\ref{tab:map_comparison}. Overall, in comparison to existing methodologies under fully supervised, semi-supervised, and sparse annotation regimes, the proposed framework demonstrates pronounced and consistent improvements in detection precision across all annotation ratios. In particular, relative to Dense Teacher \cite{DenseTeacher}, our method achieves significant gains of 11.65\%, 13.13\%, and 6.98\% under 2\%, 5\%, and 10\% label rates, respectively.
Furthermore, we conduct a comprehensive statistical analysis of the accuracy of class prompts generated by LSP module for the DOTA dataset, as presented in Table~\ref{tab:prediction_statistics}. The empirical results indicate that an increasing label rate effectively contributes to a reduction in the number of incorrect prompts. This improvement arises from the capability of leveraging annotation information embedded within the labeled data to iteratively refine the class prompts generated by the large language model. 
Additionally, as delineated in Table~\ref{tab:different_llm}, the LLM adopted in our method is specifically adapted for remote sensing, yielding superior inference performance with competitive latency compared to generic LLMs trained on conventional datasets. Furthermore, Table~\ref{tab:different_prompt} presents a comparative study on various textual prompt designs. The prompt adopted in our work was selected through extensive empirical tuning to maximize task-specific performance.

\indent \textbf{HRSC2016:} To further substantiate the efficacy of our proposed framework, we conduct comprehensive comparative analyses on the HRSC2016 dataset. The quantitative performance comparisons between our framework and various detector baselines are systematically presented in Table~\ref{tab:hrsc_comparison}. Specifically, we report the mean Average Precision (mAP) metric, as well as the AP$_{50}$ and AP$_{75}$ scores, which serve as standard benchmarks for evaluating detection accuracy. Notably, when benchmarked against the dense teacher \cite{DenseTeacher}, our framework exhibits substantial improvement of 1.94\% in terms of mAP. This empirical evidence underscores the robustness and adaptability of our approach, even when evaluated on relatively small-scale aerial image datasets. 

\subsection{Ablation Study}

\begin{figure*}[t]
  \centering
  \includegraphics[width=\textwidth]{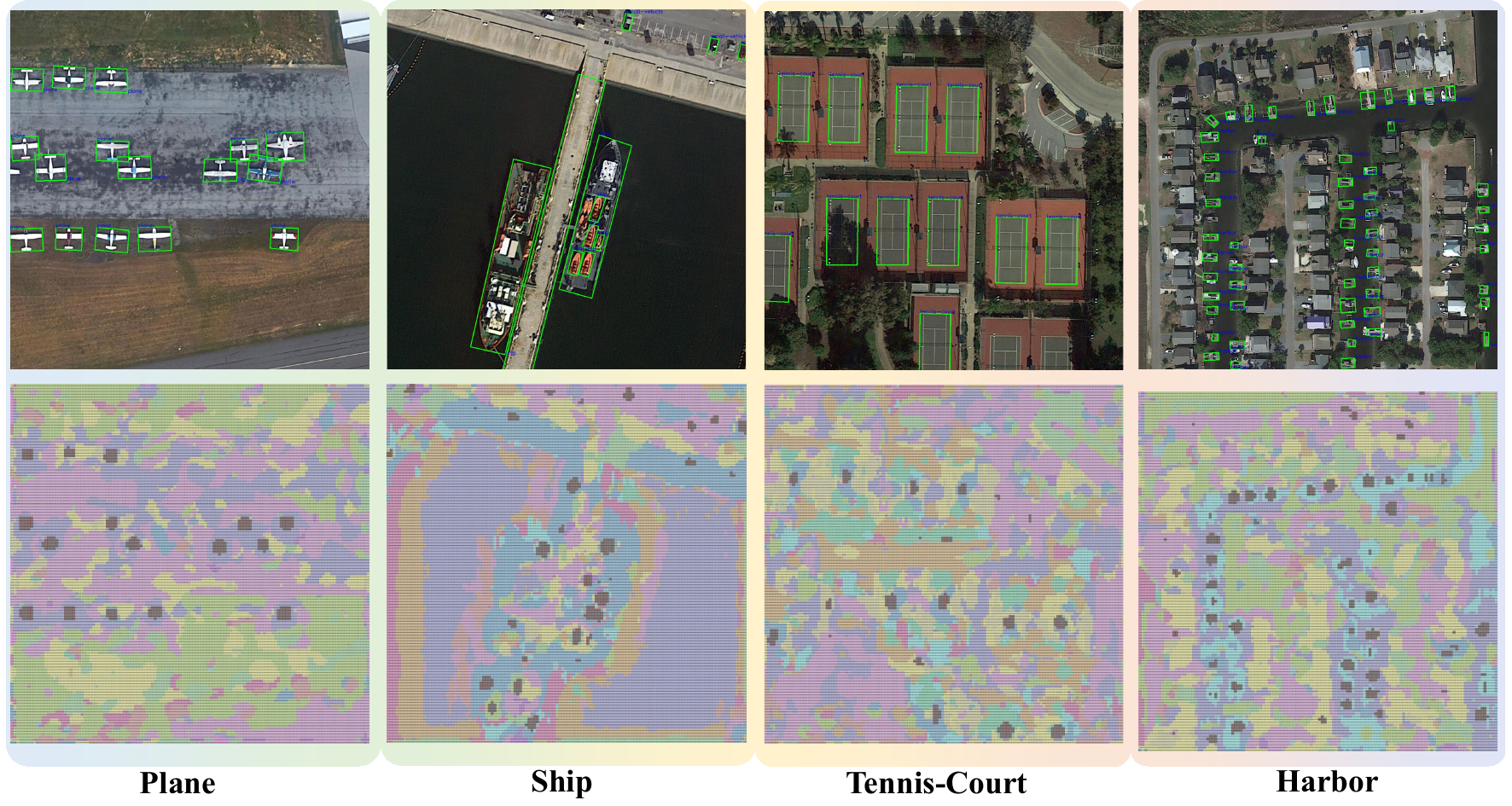}  
  \vspace{-0.6cm}
  \caption{Illustration of visualized pixel selection results with the 5\% labeling rate. The first row illustrates the original images, while the second row depicts the first-level feature maps extracted from the teacher logits. Gray dots denote the pixels identified as foreground.} 
  \vspace{-0.5cm}
  \label{fig:visual}
\end{figure*}

\begin{table}[htbp]
    \centering
    \footnotesize
    \caption{Ablation study comparing the mean Average Precision (mAP) performance resulting from the integration of various architectural components within the proposed framework. The evaluation is conducted under multiple supervision ratios (2\%, 5\%, and 10\%) to assess the effectiveness and generalization capability of each configuration.}
    \renewcommand{\arraystretch}{1.2}
    \vspace{-0.2cm}
    \begin{tabular*}{\linewidth}{@{\extracolsep{\fill}}l c c c c c c c@{}}
        \specialrule{1.2pt}{0pt}{0pt}
        & \multicolumn{4}{c}{\textbf{Components}} & \multicolumn{3}{c}{\textbf{mAP (\%) $\uparrow$}} \\
        \cmidrule(ll){2-5} \cmidrule(ll){6-8}
        & \textbf{AHR} & \textbf{LSP} & \textbf{CLA} & \textbf{MBI} & \textbf{2\%} & \textbf{5\%} & \textbf{10\%} \\
        \midrule
        \textbf{Baseline} & $\times$ & $\times$ & $\times$ & $\times$ & 47.72 & 50.82 & 58.13 \\
        \midrule
        \multirow{3}{*}{\textbf{Ours}} & $\checkmark$ & $\times$ & $\times$ & $\times$ & 55.38 & 61.60 & 62.93 \\
        & $\checkmark$ & $\checkmark$ & $\checkmark$ & $\times$ & 58.73 & 63.65 & 64.77 \\
        & $\checkmark$ & $\checkmark$ & $\checkmark$ & $\checkmark$ & 59.37 & 63.95 & 65.11 \\
        \specialrule{1.2pt}{0pt}{0pt}
    \end{tabular*}
    \label{tab:Module_Ablation}
    \vspace{-0.3cm}
\end{table}

\begin{table}[htbp]
    \centering
    \footnotesize
    \caption{Comparative Analysis of Assignment Mechanisms on the DOTA dataset under Uniform Experimental Protocols.}
    \vspace{-0.2cm}
    \label{tab:assign_strategy_comparison}
    \renewcommand{\arraystretch}{1.2}
    \setlength{\tabcolsep}{12pt}
    \begin{tabular}{lccc}
        \specialrule{1.2pt}{0pt}{0pt}
        \textbf{Assignment Strategy} & \textbf{2\%} & \textbf{5\%} & \textbf{10\%} \\
        \midrule
        SOOD                           & 54.38 & 60.02 & 63.27 \\
        Dense Teacher                  & 55.38 & 61.60 & 62.93 \\
        MCL                            & 57.63 & 62.11 & 63.27 \\
        Ours (no-prompt)     & 56.17 & 61.97 & 63.53 \\
        Ours (LLM-guided prompt)       & 58.73 & 63.65 & 64.77 \\
        Ours (gt-prompt)           & 60.45 & 65.04 & 66.41 \\
        \specialrule{1.2pt}{0pt}{0pt}
    \end{tabular}
    \vspace{-0.3cm}
\end{table}

\begin{table}[htbp]
    \centering
    \footnotesize
    \caption{Quantitative evaluation of the proposed Adaptive Hard-Negative Reweighting (AHR) module under varying configurations of weighting coefficients and prediction confidence thresholds on the training dataset.}
    \renewcommand{\arraystretch}{1.2}
    \setlength{\tabcolsep}{20pt}  
    \vspace{-0.2cm}
    \begin{tabular}{c c c c}
        \specialrule{1.2pt}{0pt}{0pt}
        \multirow{2}{*}{\textbf{Weight}} & \multicolumn{3}{c}{\textbf{Threshold}} \\
        \cmidrule(ll){2-4}
        & \textbf{1.0} & \textbf{0.95} & \textbf{0.90} \\
        \midrule
        0.30 & 67.87 & 67.53 & 67.76 \\
        0.25 & 66.82 & 67.81 & 67.70 \\
        0.20 & 67.89 & 68.01 & 67.90 \\
        0.15 & 67.31 & 67.69 & \textbf{68.20} \\
        0.10 & 67.83 & 67.73 & 68.01 \\
        \specialrule{1.2pt}{0pt}{0pt}
    \end{tabular}
    \label{tab:AHR_Ablation}
    \vspace{-0.3cm}
\end{table}

\begin{figure}[t]
  \centering
  \includegraphics[width=0.475\textwidth]{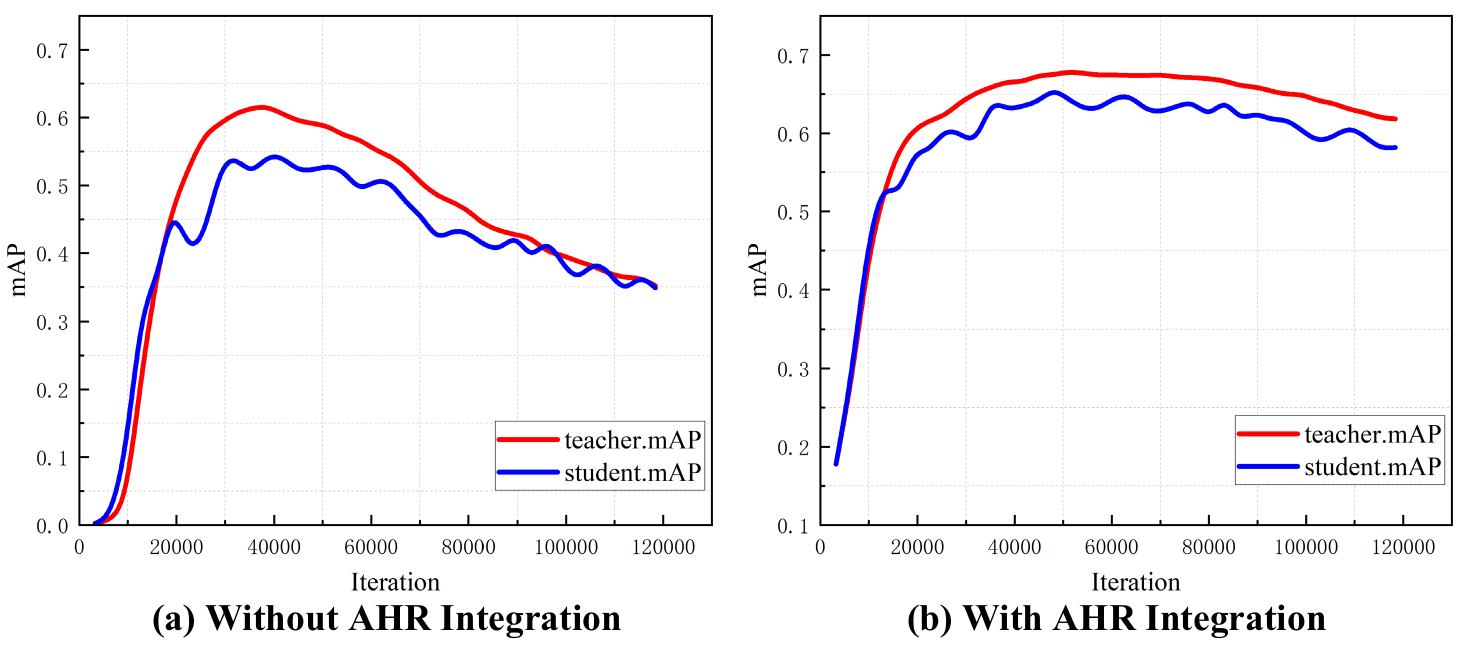} 
  \vspace{-0.6cm}
  \caption{Impact of the AHR Integration on the mAP Performance.}
  \label{fig:mAP} 
  \vspace{-0.6cm}
\end{figure}

\indent We conduct comprehensive ablation experiments to evaluate the performance of our framework under different settings. All the experiments are conducted  on the DOTA dataset at the 5\% label rate, unless stated otherwise.

\indent As illustrated in Table~\ref{tab:Module_Ablation}, we conduct a rigorous ablation study to systematically evaluate the contribution of each proposed component. The experimental findings reveal that even when solely employing the AHR, our approach already surpasses the baseline, highlighting the effectiveness of adaptive negative sample  in improving detection performance. Furthermore, the incorporation of the LSP, CLA and MBI modules provides additional performance gains, substantiating their respective roles in enhancing feature representation learning and refining pseudo-label assignments. These results collectively underscore the efficacy of our proposed framework in leveraging complementary strategies to achieve superior detection accuracy under limited supervision. 
Furthermore, Fig.~\ref{fig:visual} provides a visualization of the foreground pixels selected by our model during training. Notably, the distribution of the identified positive samples aligns closely with the locations of ground truth bounding boxes in the original images, further validating the effectiveness of our assignment strategy in accurately capturing salient object regions.

\indent To address the premise that erroneous predictions may impart a non-trivial perturbation to convergence dynamics, our assignment framework integrates three foundational mechanisms deliberately formulated to suppress such detrimental influences. As depicted in Table~\ref{tab:assign_strategy_comparison}, we conducted empirical assessments under both the ``no-prompt'' and ``gt-prompt'' configurations, which correspondingly delineate the upper-bound performance and facilitate systematic benchmarking against alternative assignment paradigms.

\indent The impact of hyperparameter selection within AHR is systematically analyzed, with the corresponding experimental results summarized in Table~\ref{tab:AHR_Ablation}. In this study, we investigate the effects of two key hyperparameters: the threshold parameter $thr$, which determines the selection criterion for hard negative samples, and the weighting factor $w$, which governs the degree of down-weighting. Our empirical findings indicate that while the overall performance remains relatively stable across different hyperparameter configurations, 
the optimal detection performance of 68.20\% on the training set is achieved  when the threshold $thr$ is set to 0.9 and the weight $w$ to 0.15.
Moreover, Fig.~\ref{fig:mAP} presents the comparative analysis of mAP curves before and after applying AHR. A noticeable improvement in the stability of mAP progression is observed, wherein the curve exhibits a smoother convergence trend, eliminating the abrupt declines that are typically present in the absence of AHR. This empirically substantiates the effectiveness of our strategy in mitigating instability caused by noisy negative samples, ultimately leading to more reliable and robust learning dynamics throughout the training process.

\vspace{-0.2cm}\section{Conclusion}

\indent This paper introduces an LLM-assisted semantic guidance framework based on a Multi-Branch Input architecture to enhance pseudo-label reliability for sparsely annotated object detection in remote sensing imagery. The Adaptive Hard-Negative Reweighting module alleviates optimization biases in the supervised branch, while the LLM-Assisted Semantic Prediction module refines class-specific prompts to facilitate Class-Aware Label Assignment. Extensive experiments on public benchmarks demonstrate the effectiveness of our approach in mitigating category imbalance and assignment ambiguities, achieving superior performance over existing methods. Future work will explore the broader applicability and adaptation of our method to varying levels of label sparsity.

\vspace{-0.2cm}\section*{Acknowledgements}

\indent This work is supported by the National Natural Science Foundation of China (Grants Nos. 62372238, 62476133).
{
    \small
    \bibliographystyle{ieeenat_fullname}
    \bibliography{main}
}

\end{document}